\documentclass[a4paper]{llncs}

\usepackage{graphicx,url}
\usepackage[T1]{fontenc}
\usepackage[american]{babel}

\usepackage{color} 

\title{%
Visual Inference Specification Methods\\ for Modularized Rulebases.\\ Overview and Integration Proposal%
\thanks{The paper is supported by the \textit{BIMLOQ} Project funded from 2010--2012 resources for science as a research project.}}
\author{Krzysztof Kluza, Grzegorz J. Nalepa, \L{}ukasz \L{}ysik}
\institute{Institute of Automatics, \\
  AGH University of Science and Technology,\\
  Al. Mickiewicza 30, 30-059 Krak\'ow, Poland\\
\email{\texttt{kluza@agh.edu.pl,gjn@agh.edu.pl}}
}

\begin{document}
\bibliographystyle{splncs}
\maketitle

\begin{abstract}
The paper concerns selected rule modularization techniques. Three visual methods for inference specification for modularized rulebases are described: Drools Flow, BPMN and XTT2. Drools Flow is a~popular technology for workflow or process modeling, BPMN is an~OMG standard for modeling business processes, and XTT2 is a~hierarchical tabular system specification method. Because of some limitations of these solutions, several proposals of their integration 
are given.
\end{abstract}

\section{Introduction}
\label{sec:intro}

Rule-Based Systems (RBS)~\cite{ali-book-springer} constitute one of the most powerful knowledge representation formalisms.
Rules are intuitive and easy to understand for humans.
Therefore, this approach is suitable for many kinds of computer systems.
Nowadays, software complexity is increasing. 
Because describing it using plain  text is very difficult, 
many visual methods have been proposed.
For example, in Software Engineering, the dominant graphical notation for software modeling is UML (the Unified Modeling Language).
Design of large knowledge bases is non trivial, either.
However, in the area of RBS, there is no single visual notation or coherent method. 
Moreover, the existing solutions have certain limitations.
These limitations are especially visible in the design of large systems.



When the number of rules grows, system scalability and maintainability suffers.
To avoid this, there is a need to manage rules.
Rule grouping is a simple method of rule management.
However, it is not obvious how to group rules. 
One of the most common grouping methods ivolves context awareness and creation of decision tables.
Another grouping method takes rule  dependencies into account and creates decision trees.
This leads to RBS modularization. 

This paper describes three possible solutions to modularize rule bases: 
\begin{itemize}
\item Drools Flow~\cite{browne2009drools}, which is a popular technology for workflow 
modeling,
\item BPMN (the Business Process Modeling Notation)~\cite{owen:raj:bpmn}, which is an OMG standard~\cite{omg:2006:bpmn1.0} for modeling business processes, and
\item XTT2 (EXtended Tabular Trees), which is a result of authors' research project~\cite{gjn2009amcs} and which organizes a~tabular system into a hierarchical 
structure.
\end{itemize}

However, these solutions have some limitations. 
Drools Flow is platform-dependent and not standarized. 
Moreover, it has some flow design restrictions.
BPMN is a notation for business processes, and it is not clearly stated how processes can co-operate with rules. 
Furthermore, BPMN can be mapped to BPEL (Business Process Execution Language) for execution, but this mapping is non-trivial and execution is not possible for every 
BPMN model.
XTT2, in turn, is not wide-spread, and it is not a universal method. 


The general problem considered in this paper is the RBS design and modularization.
The article constitutes an overview and a proposal for integration of the three presented methodologies, 
which can be useful in solving above-mentioned problems.
The next two sections present selected rule modularization techniques 
and an overview of selected visual design methods for rule inference. 
In Section~\ref{sec:xtt2drools}, a~proposal of rule translation from XTT2 to Drools is described, 
and in Section~\ref{sec:bpmn4xtt2}, a~proposal of XTT2 inference design with BPMN is introduced. 
Section~\ref{sec:future}, discusses future work and 
summarizes the main threads of this article. 

\section{Rule Modularization Techniques}
\label{sec:mods}



Most  
classic expert systems have a flat knowledge base.
So, the inference mechanism has to check each rule against each fact.
When the knowledge base is large, 
this process becomes inefficient.
This problem can be solved by providing a structure in the knowledge base that allows to only check a subset of rules~\cite{bobek2010ict}.

CLIPS~\cite{Giarratano:2005} allows for  organising rules into so-called \textit{modules}, that restrict access to their elements from other modules. 
Modularisation of the knowledge base helps rule management. 
In CLIPS, each module has its own pattern matching network for its rules and its own agenda. 
Execution focus can be changed between modules stored on the 
stack.

JESS~\cite{friedman-hill2003:jess} also provides a module mechanism. 
Modules provide structure and 
control execution. 
In general, although any JESS rule can be activated at any time, only rules in the focus module will fire.
This leads to a structured rule base, but still all rules are checked against the facts.
In terms of efficiency, the module mechanism does not influence on the performance of \textit{conflict set} creation.

Drools Flow 
provides a graphical interface for modelling of processes and rules. 
Drools 5 has a built-in functionality to define the structure of the rule base, which can determine the order of rule evaluation and execution.
Rules can be grouped in ruleflow-groups which define the subsets of rules that are 
executed.
The ruleflow-groups have a graphical representation as nodes on the \textit{ruleflow} diagram. 
They are connected with links, which determines the order of evaluation.
Rule grouping in Drools 5 contributes to the efficiency of the ReteOO algorithm, because only a subset of rules is evaluated. 
However, there are no policies which determine when a rule can be added to the ruleflow-group.

\section{Selected Visual Design Methods for Rule Inference}
\label{sec:vis}

Efficient inference would not be possible without a~proper structure and design of rule-based system.
The important issues in dealing with this problem is grouping and hierarchization of rules, as well as addressing the contextual nature of the~rulebase.
The following subsections describe selected methods and tools, in which visual design of rule inference is possible.

\subsection{Drools}
\label{sec:drools}

Drools is a rule engine which offers knowledge integration mechanisms. The~project is run by the JBoss Community, which belongs to the Red Hat Foundation. It is divided into four subprojects: Guvnor, Expert, Flow and Fusion. Each of them supports different part of integration process.


Expert is the essence of Drools. It is the actual rule engine. It collects facts from the environment, loads the knowledge base, prepares the agenda and executes rules. A modified version of the Rete~\cite{ProductionMatching} algorithm is used for 
the inference. 





The \emph{knowledge base} in Drools consists of three main elements: rules, decision tables and Drools Flow. 
%
%
The fundamental form of knowledge representation in Drools is a rule. 
This form is easy to use and very flexible. Rules are stored in text files which are loaded into the program memory by special Java classes.
Rules in Drools can be suplemented with attributes which contain additional information. They have a form of name-value pairs and they describe such paramters as rule priority and provide meta information for inference engine.

Rules which have the same schema can be combined into \textit{decision tables}. A~decision table is devided into two parts: the \textit{left-hand side}, which represents the conditions of rules and the \textit{right-hand side}, which represent the actions to be executed. One row in a table corresponds to one rule. 
However, decision tables, are useful only during the design phase. The structure does not improve the~performance of the inference. Decision tables are, in fact, transformed into rules. So the inference engine does not recognize which rules come from decision tables and which are just a group of unrelated rules.

\begin{figure}[ht]
\vspace{-3mm}
\centering
\includegraphics[width=0.6\textwidth]{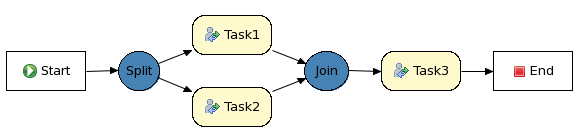}
\caption{Sample Drools Flow diagram (drools.org)}
\label{fig:sampleDroolsFlow}
\vspace{-3mm}
\end{figure}

Rules form a flat structure. When the inference engine matches rules against facts, it takes all rules into consideration. The user, however, can define the flow of the inference process. Drools Flow offers a workflow design functionality in the form of blocks (See Fig.~\ref{fig:sampleDroolsFlow}). The user can specify exactly which rules should be executed in which order and under which conditions. 

Each model in Drools Flow has to contain two blocks: \textit{start} and \textit{end}. Rules and rule flow are linked together inside the \textit{ruleset} block. Each \textit{ruleset} block has a \textit{ruleflow-group} attribute. Similarly, each rule has the attribute with the same name. Rules belong to the ruleset block with the same values of the ruleflow-group attribute.
Additionally, the process can be split and joined. Two blocks, \textit{split} and \textit{join}, are used for that purpose. The block \textit{split} has different types. The~AND type defines that the process follows all the outgoing connections. 
The~\textit{join} block also has different types. The AND type waits for all the incomming subprocesses to finish. The OR type waits for the first process to finish, while the n-of-m type waits until specified number of processes finish.

Drools has some \emph{limitations}. First of all, it is not a standarized solution. The~form of knowledge representation still evolves. It could have been seen when version 5 was released - new block were introduced and the format of Drools Flow file has changed.
Moreover, Drools does not provide any tools which can be used in the knowledge design phase. It can be problematic in large systems. What is more, the rulebase has a flat structure. Although, Drools Flow complements the~strucutre by desribing execution process, the rules still do not have a hierarchy.
The last thing is that Drools is language dependent, 
closely related to 
Java. 
Parts of the rules and some Rule Flow blocks  contain Java expresions. 

\subsection{XTT2}
\label{sec:xtt}

XTT2 (EXtended Tabular Trees)~\cite{gjn2009amcs} is a hybrid knowledge representation and design method 
aimed at combining decision trees and decision tables.
It has been developed in the HeKatE research project (\url{hekate.ia.agh.edu.pl}), 
and its goal is to provide a new software development methodology, 
which tries to incorporate some well-established Knowledge Engineering tools and paradigms into the domain of Software Engineering, such as declarative knowledge representation, knowledge transformation based on existing inference strategies as well as verification, validation and refinement. 

\begin{figure}[ht]
\vspace{-5mm}
\centering
\includegraphics[width=0.9\textwidth]{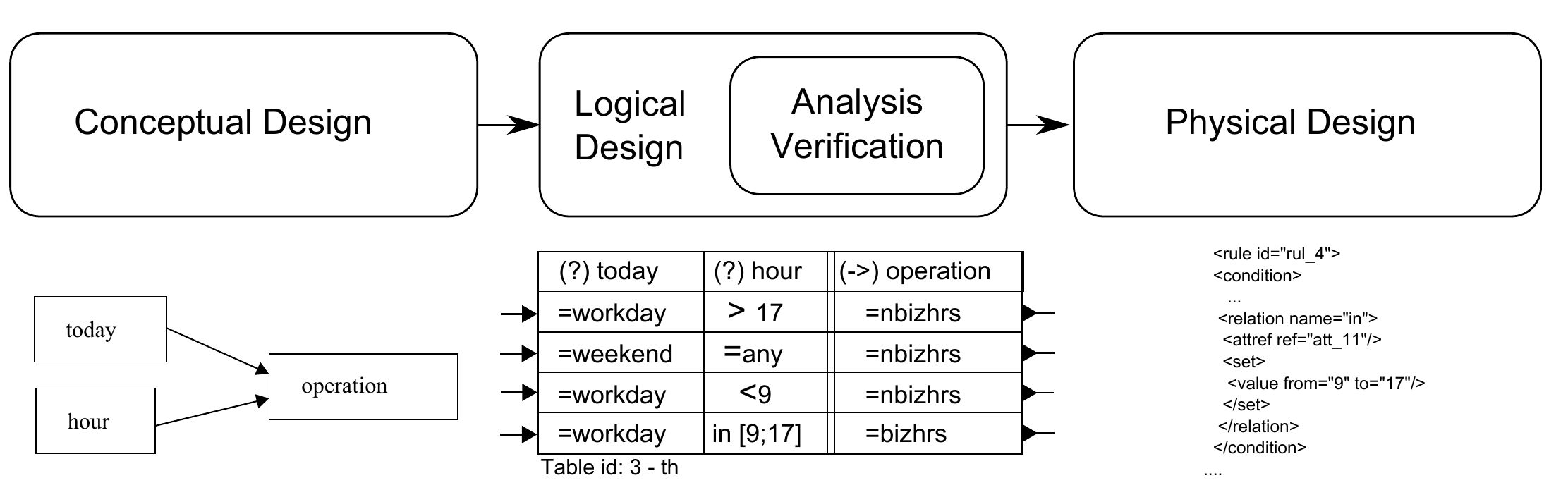}
\caption{The HeKatE process}
\label{fig:HeKatEprocess}
\vspace{-3mm}
\end{figure}

The HeKatE process consists of three design phases (shown in Fig.~\ref{fig:HeKatEprocess})~\cite{gjn2009amcs}:
\begin{enumerate}
\item \textbf{The conceptual design phase}, which is the most abstract phase. During this phase, both system attributes and their functional relationships are identified. This phase uses ARD+ (Attribute-Relationship) diagrams as a~modeling tool. It allows design of the logical XTT2 structure.
\item \textbf{The logical design phase}, in which system structure is represented as a~XTT2 hierarchy. The preliminary model of XTT2 can be obtained as a~result of the previous phase. This phase uses the XTT2 representation as a~design tool. During this phase, on-line analysis, verification as well as revision and optimization (if necessary) of the designed system properties 
is provided.
\item \textbf{The physical design phase}, in which 
the system implementation is generated from the XTT2 model. 
The 
code can be executed and debugged.
\end{enumerate}
Some \emph{limitations} of XTT2 can be pointed out.
XTT2 provides a support for the~entire process. It is used to model, represent, and store the business logic of designed systems. 
Rules in XTT2 are formalized with the use of the ALSV(FD)~\cite{gjn2009amcs} logic and are supported by a Prolog-based interpretation.
Although XTT2 rules are prototyped with the ARD+ method, the method is quite poor, and does not provide more advanced workflow constructs.
Moreover, it is not a widely known methodology and only dedicated tools support it. 


\subsection{Business Rules and BPMN}
\label{sec:bpmn}



\begin{figure}[ht]
\vspace{-2mm}
\centering
\includegraphics[scale=0.29]{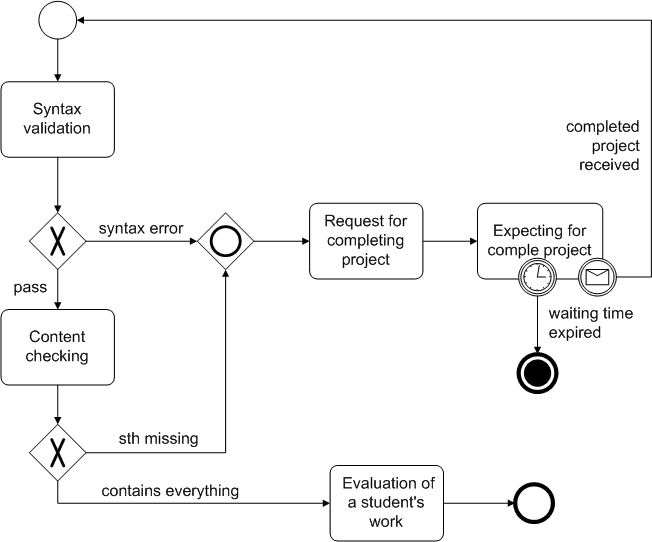}
\caption{An example of Business Process Diagram}
\label{fig:bpmnExample}
\vspace{-4mm}
\end{figure}

BPMN~\cite{omg:2006:bpmn1.0} is a visual notation for business processes.
A BPMN model defines the~ways in which operations are carried out to accomplish the intended objectives of an organization. 
Visualization makes the model easier to understand.
The goal of the notation is to provide such a notation 
which is easily understandable by business users.
The notation provides only one kind of diagram~-- BPD (Business Process Diagram).
There are four basic categories of BPD elements: Flow Objects (Events, Activities, and Gateways), Connecting Objects (Sequence Flow, Message Flow, Association), Swimlanes, and Artifacts.
An example describing evaluation process of a student project is presented in Fig.~\ref{fig:bpmnExample}.


BPMN~\cite{omg:2006:bpmn1.0} has been developed by the Business Process Management Initiative (BPMI) and currently is maintained by the Object Management Group.
Although the notation is relatively young, BPMN is becoming increasingly popular. 
According to OMG, there is more than 60 BPMN implementations of BPMN tools. 
Moreover, BPMN models can be serialized to XML and further processed e.g. into languages for execution of business processes, 
such as BPEL4WS (Business Process Execution Language for Web Services)~\cite{sarang2006}.



\begin{figure}[ht]
\vspace{-2mm}
\centering
\includegraphics[scale=0.4]{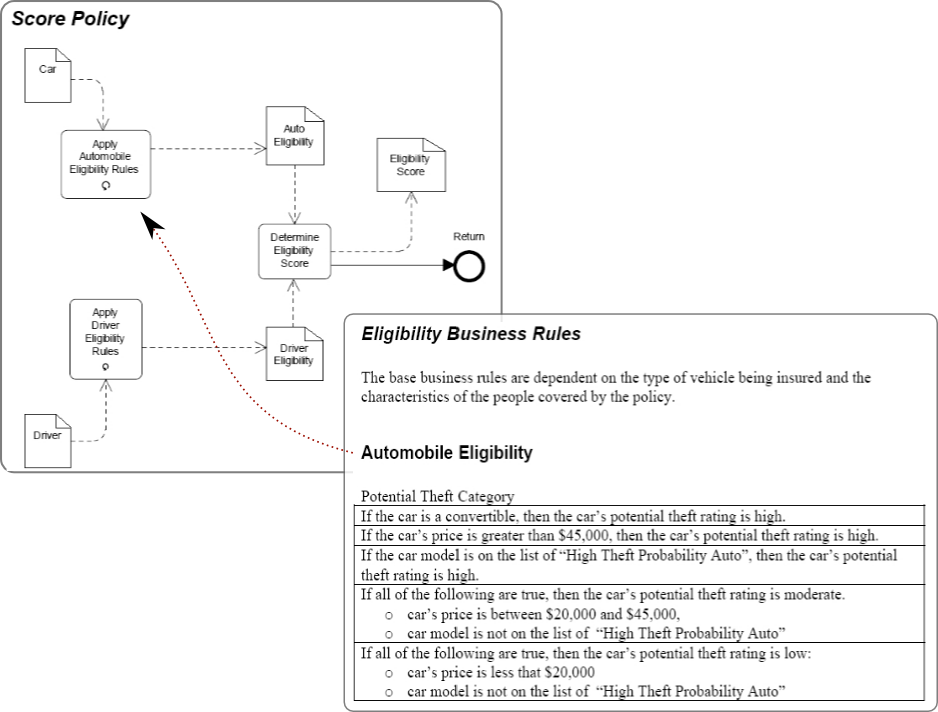}
\caption{An example of using BR to define a process in Business Process Diagram}
\label{fig:brvsbp}
\vspace{-2mm}
\end{figure}

Very often a Business Process (BP) is associated with particular Business Rules (BR), which define or constrain some business aspect, and are intended to assert business structure or to control or influence the business behavior~\cite{hay:kolber:2000:br}. 
According to the specification~\cite{omg:2006:bpmn1.0}, BPMN is not suitable for modeling concepts, such as organizational structures and resources, data models, and business rules.
There is a~huge difference in abstraction level between BPMN and~BR. 
However, BR may be complementary to the business process. 
In~Fig.~\ref{fig:brvsbp}, an~example from the classic UServ Financial Services case study 
has been shown.
This example presents how business processes and rules can be linked.

BPMN has some \emph{weaknesses}. 
Although a specification defines a~mapping between BPMN and BPEL (standard for execution languages), there is a~fundamental difference between these two standards.
One of the consequences of this difference is that, for instance, not every BPMN process can be mapped to BPEL and executed.
Moreover, execution of the processes requires additional specification, which is not necessarily integrated with the entire design process.

Despite the fact that the BPMN model has well-defined semantics and a~particular model should be clearly understood, 
there can be various models having the same meaning 
and there can be ambiguity in sharing BPMN models.
Last but not least, it is difficult to asses the quality of the model. 


\subsection{Critical comparison}
\label{sec:critics}

As one can see from the Table~\ref{tab:comparison}, each of these solutions has some pros and cons. 
Integration of these technologies based on their merits can bring better results than using them separately.

\begin{table}[ht]
\vspace{-1mm}
\begin{center}
\begin{tabular}{|l|l|l|l|} \hline
\textbf{~} & \textbf{Drools~4} & \textbf{BPMN} & \textbf{XTT2} \\ \hline \hline
\textbf{Visual design of the rulebase} & no & no & \textbf{yes} \\ \hline
\textbf{Verification} & no & some  & \textbf{yes} \\ \hline
\textbf{Workflow modeling (OR, AND etc.)} & yes & \textbf{yes} & no \\ \hline
\textbf{Runtime environment} & yes & no & \textbf{yes} \\ \hline
\textbf{Tool support} & yes & \textbf{yes} & yes \\ \hline
\textbf{Standardization} & no & \textbf{yes} & no \\ \hline
\end{tabular}
\end{center}
\caption{Comparison of the three approaches}
\label{tab:comparison}
\vspace{-7mm}
\end{table}

The disadvantages of the Drools Flow are platform dependency and lack of standarization.
Drools Flow supports decision tables and grouping of unrelated rules. 
XTT2 allows multiple connections between tables.
Although Drools only allows for a single connection, it provides \textit{Join} and \textit{Split} blocks.

The XTT2 connections are of the AND type, by default.
However, the conection semantics is different than that in Drools or BPMN.
In Drools and BPMN, the default inference process is forward chaining, 
while XTT2 provides various inference modes, e.g. forward chaining (where the connections ore of the AND type) or backward chaining (where the semantics of connections varies). 

BPMN is only a notation which has many elements for precise control of flow. 
However, this solution originally was not based on Rule-Based Systems.
Therefore, it does not define the relationship between processes and rules.
Although BPMN can be mapped to BPEL and executed, 
mapping and execution is possible only for selected groups of a BPMN model. 

In case of XTT2, the entire design process is supported.
What is more, formal on-line analysis can be performed during the design process,
and then a~prototype of the~system can be generated.
However, XTT2 is not a~wide-spread solution, and does not pretend to be a~universal method.

On the one hand, the comparison shows that XTT2 is the only one solution which supports visual modeling of the rulebase 
(modeling using decision tables). Moreover, only XTT2 provides formal verification.
On the other hand, Drools offers workflow modeling. 
The integration of Drools and XTT gave the opportunity to combine these advantages.
The next section describes the proposal of rule translation from XTT2 to Drools in detail,
as part of the HeKatE project.

BPMN is already a well-known and standardized notation. 
In Drools~5, it can be used to model workflow.
To facilitate workflow modeling for XTT2 and to provide an executable platform for BPMN, the integration of XTT2 with BPMN is considered. 
The possible scenarios are identified and described in Section~\ref{sec:bpmn4xtt2}.
This research is a part of the \textit{BIMLOQ} project (2010--2012).

\section{Proposal of Rule Translation from XTT2 to Drools}
\label{sec:xtt2drools}


Knowledge structure represented by Drools is very similar to the one represented by XTT2. In fact, that was one of the main reasons for choosing Drools as an integration platform. Both frameworks have the same goal: to provide rule-based and structurized knowledge representation. On the one hand, XTT2 is a unified structure which contains both rules and inference flow. On the other hand, Drools has both of these features, but rules can exist without Drools Flow.
Both solutions can be used to model  business processes. Drools Flow even provides special blocks which contain Java source code to be executed. XTT2, however, is more flexible and language independent. It contains rules which do not have any dialect specific parts.

\subsection{Generating Drools files}
\label{sec:genDroolsFiles}

Knowledge represented in XTT2 is stored in XML form. One file contains a~tree structure and rules. Drools with the Flow model, on the other hand, stores knowledge in at least two files: a file with rules and a file with a flow. The~XTT2-to-Drools integration mechanism separates XTT2 rules from the structure, transforms them and puts into two separate Drools files.

Nevertheless, Drools operates on objects while XTT2 uses primitive types. In Drools, facts are instances of Java classes inserted into the working memory. When the rules are fired, values used during comparison are taken from  objects using getters. The workaround would be to create one Java class which contains all XTT2 attributes. The class is called a \textit{Workspace}.
To sum up, three files are generated from one XTT2 model file: \textit{Rule Flow} (model structure), \textit{Decision tables} (aggregated rules), and \textit{Workspace} (a Java class with all  attributes).

The results of XTT2 into Drools translation are three files. 
The first one is an XML based file and represents the flow structure. It does not contain the~actual rules, but only the nodes (tables' names). The second one, a~CSV (Comma separated values) file, contains Decision Tables storing the rules. The last one is a single Java class which holds all the XTT2 attributes. 


\subsection{Structural difference}

While generating Drools files from an XTT2 file, structural differences are revealed. First one was already mentioned above. It is the form of attribute types. There is, however, an easy solution. The type of every XTT2 attribute is exchanged with an appropriate Java type. All XTT2 attributes are wrapped into one \textit{Workspace} class which 
does not contain any logic but getters and setters.

Another structural difference is the placement of the logical operator. An~XTT2 table is translated to a Drools decision table. An XTT2 table contains logical operators in the table cell~-- together with the value used in comparison. This implies that in one column, many different operators can appear. In Drools, however, the logical operator is placed in a table header. This means that all cells underneath use the same operator. This problem can be solved by decomposing the XTT2 columns into one or more columns in Drools model. Table~\ref{tab:xttTableFromThermostatEx} is the~representation of XTT rules, while Table \ref{tab:decTabForThermostatEx} is its Drools equivalent.

\begin{table}[ht]
\vspace{-2mm}
\begin{center}
\begin{tabular}{|l|l|l|} \hline
\textbf{today} & \textbf{hour} & \textbf{operation} \\ \hline \hline
= workday & > 17 & = nbizhrs \\ \hline
= weekend & = ANY & = nbizhrs \\ \hline
= workday & < 9 & = nbizhrs \\ \hline
= workday & in [9,17] & = bizhrs \\ \hline
\end{tabular}
\end{center}
\caption{XTT table from the thermostat example}
\label{tab:xttTableFromThermostatEx}
\vspace{-9mm}
\end{table}

\begin{table}[ht]
\vspace{-5mm}
\begin{center}
\begin{tabular}{|p{0.15\textwidth}|p{0.15\textwidth}|p{0.15\textwidth}|p{0.15\textwidth}|p{0.15\textwidth}||p{0.15\textwidth}|} \hline
\footnotesize condition & \footnotesize condition & \footnotesize condition & \footnotesize condition & \footnotesize condition & \footnotesize action \\ \hline \hline
\footnotesize Workspace & \footnotesize Workspace & \footnotesize Workspace & \footnotesize Workspace & \footnotesize Workspace & \footnotesize Workspace \\ \hline
\footnotesize Today = ''\$param'' & \footnotesize hour > & \footnotesize hour < & \footnotesize hour >= & \footnotesize hour <= & \footnotesize setOperation (''\$param'') \\ \hline \hline
workday & 17 & & & & nbizhrs \\ \hline
weekend & & & & & nbizhrs \\ \hline
workday & & 9 & & & nbizhrs \\ \hline
workday & & & 9 & 17 & bizhrs \\ \hline
\end{tabular}
\end{center}
\caption{Decision Table for the thermostat example}
\label{tab:decTabForThermostatEx}
\vspace{-7mm}
\end{table}


There are some structural differences in the flow structure as well. First of all, XTT2 tables allow multiple incoming connections. Furthermore, the connection can be directed to a~specific row in a~table. It is not possible in Drools Flow.  \textit{Ruleset} blocks can have only one incoming and one outgoing connection. This issue can be resolved by placing \textit{split} and \textit{join} blocks before and after the \textit{ruleset} block. Nevertheless, the problem with row-to-row connection is still present and it is to be resolved in a future version of the integration proposal.

Another difference with the representation form of Drools Flow appeared when version 5 of Drools was released. In version 4.0.7, the Drools Flow structure could only be created in a dedicated Eclipse plugin. This is because the file which contained the structure was a~Java class serialized using the XStream library (\url{http://xstream.codehaus.org}). The programmer was dependent on the class, which was included into the plugin. In version~5 however, the file storing the Flow structure was slimmed and now contains only the most important information: blocks defined in the flow and connections between them. 

\section{Proposal of XTT2 Inference Design with BPMN}
\label{sec:bpmn4xtt2}

The integration of XTT2 with BPMN faces two main challenges. 

\textbf{Different goals}:
BPMN provides a notation for modeling business processes. 
Such processes define the order of tasks to accomplish the intended objectives of an organization. 
Although in BPMN one can define very detailed description of the particular task, 
it is rather not the proper use of the notation.
The XTT2 methodology, in turn, is not only a notation. 
It provides well-founded systematic and complete design process~\cite{gjn2009amcs}. 
This preserves the quality aspects of the rule model 
and allows gradual system design
and automated implementation of RBS. 

\textbf{Different semantics}
Apart from goals, the semantics of both notations is also different. 
BPMN describes processes while XTT2 provides the description of rules. 
Although the semantics of each BPMN element is defined, 
the implementation of some particular task is not defined in pure BPMN.
XTT2 provides a formal language definition and therefore enables automatic verification and execution.
Therefore, BPMN and XTT2 operate on different abstraction levels. 

\begin{figure}[ht]
\vspace{-2mm}
\centering
\includegraphics[scale=0.3]{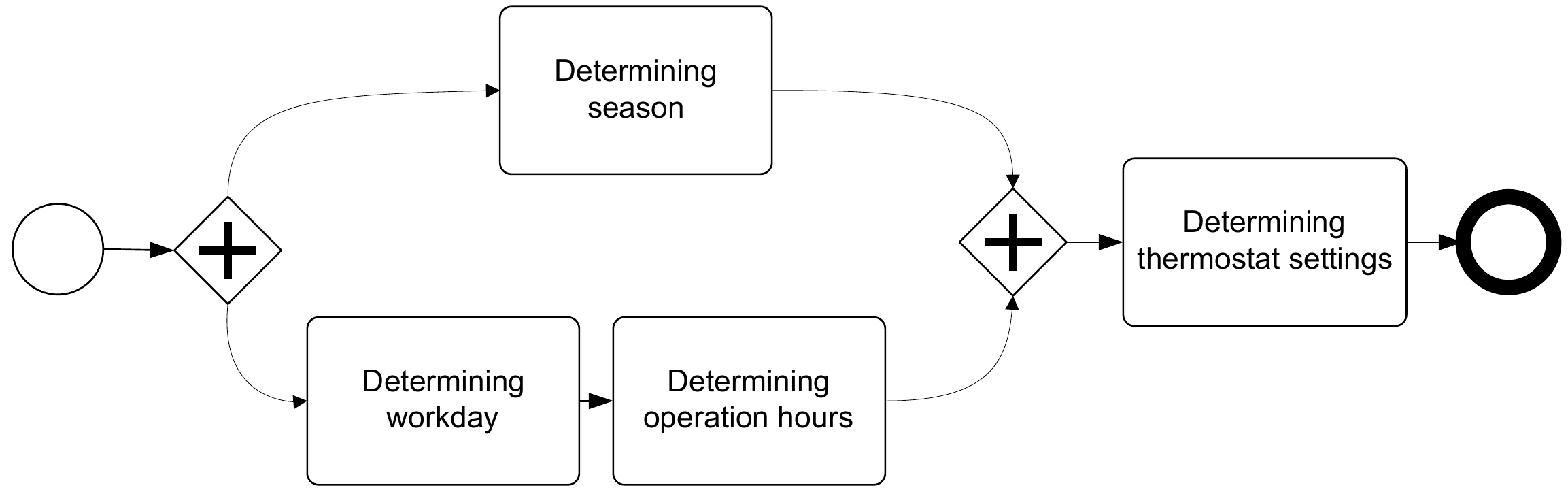}
\caption{An example of using BPMN instead of ARD+}
\label{fig:visio_therm}
\vspace{0mm}
\end{figure}

\begin{figure}[ht]
\vspace{0mm}
\centering
\includegraphics[scale=0.55]{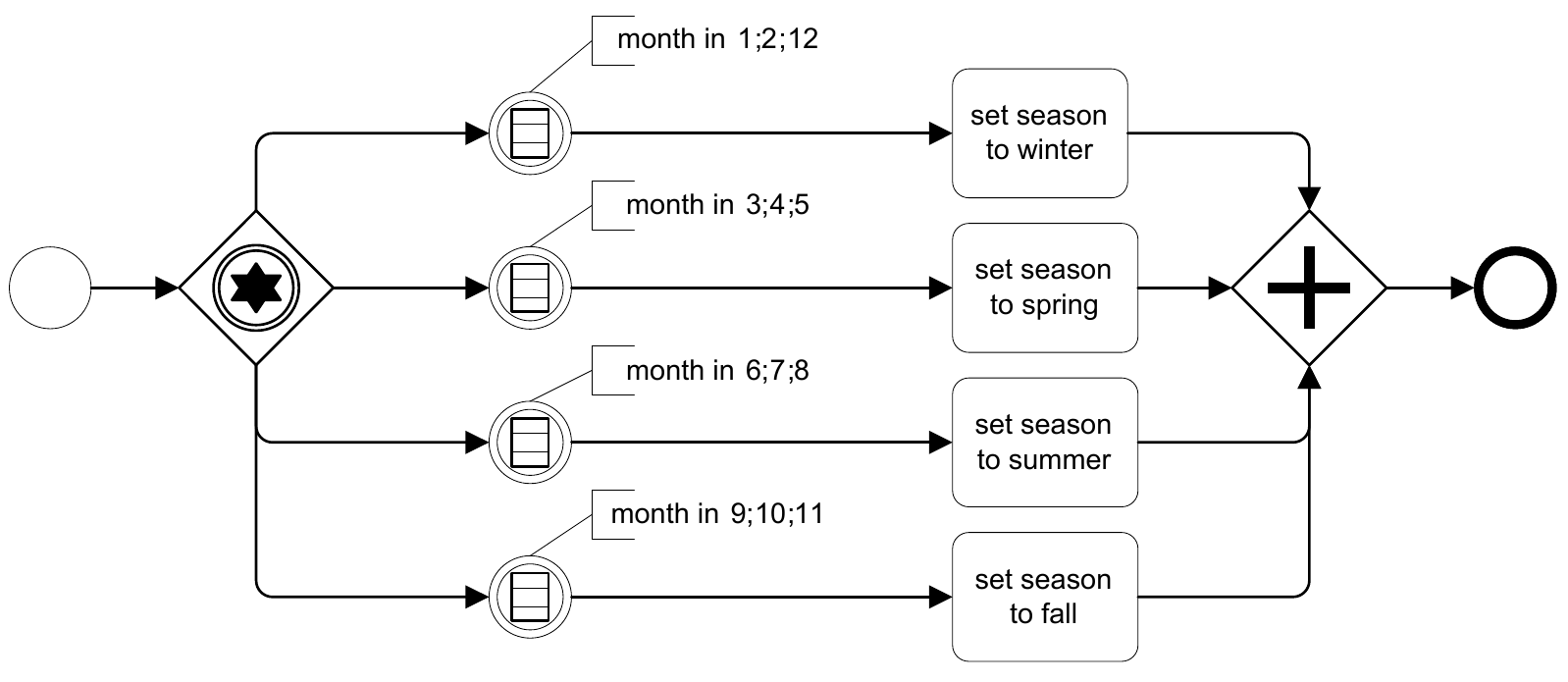}
\caption{BPMN representation of XTT2 table}
\label{fig:tab_ms_bpmn}
\vspace{-4mm}
\end{figure}

\begin{figure}[ht]
\vspace{-3mm}
\centering
\includegraphics[scale=0.45]{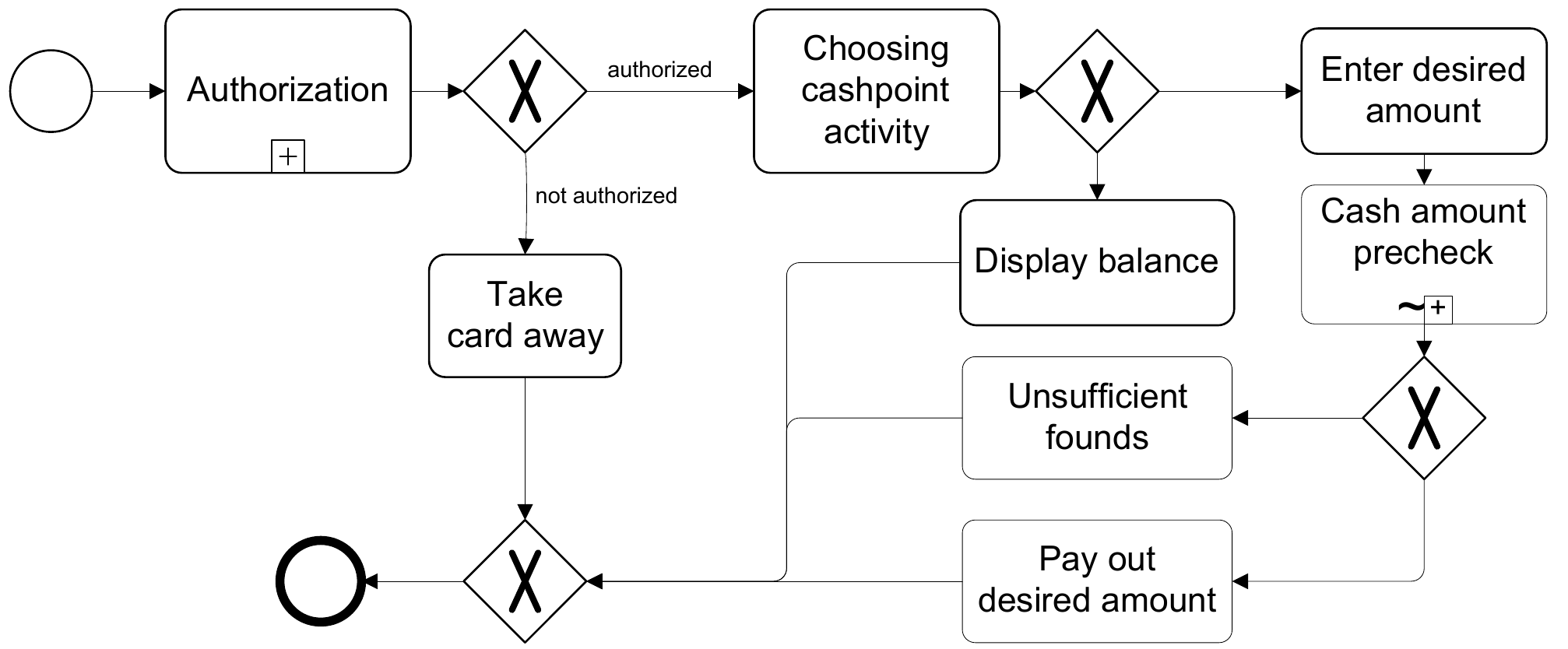}
\caption{Example of BPMN representation of cashpoint}
\label{fig:cashpoint_diagram}
\vspace{0mm}
\end{figure}

\begin{figure}[ht]
\vspace{0mm}
\centering
\includegraphics[scale=0.35]{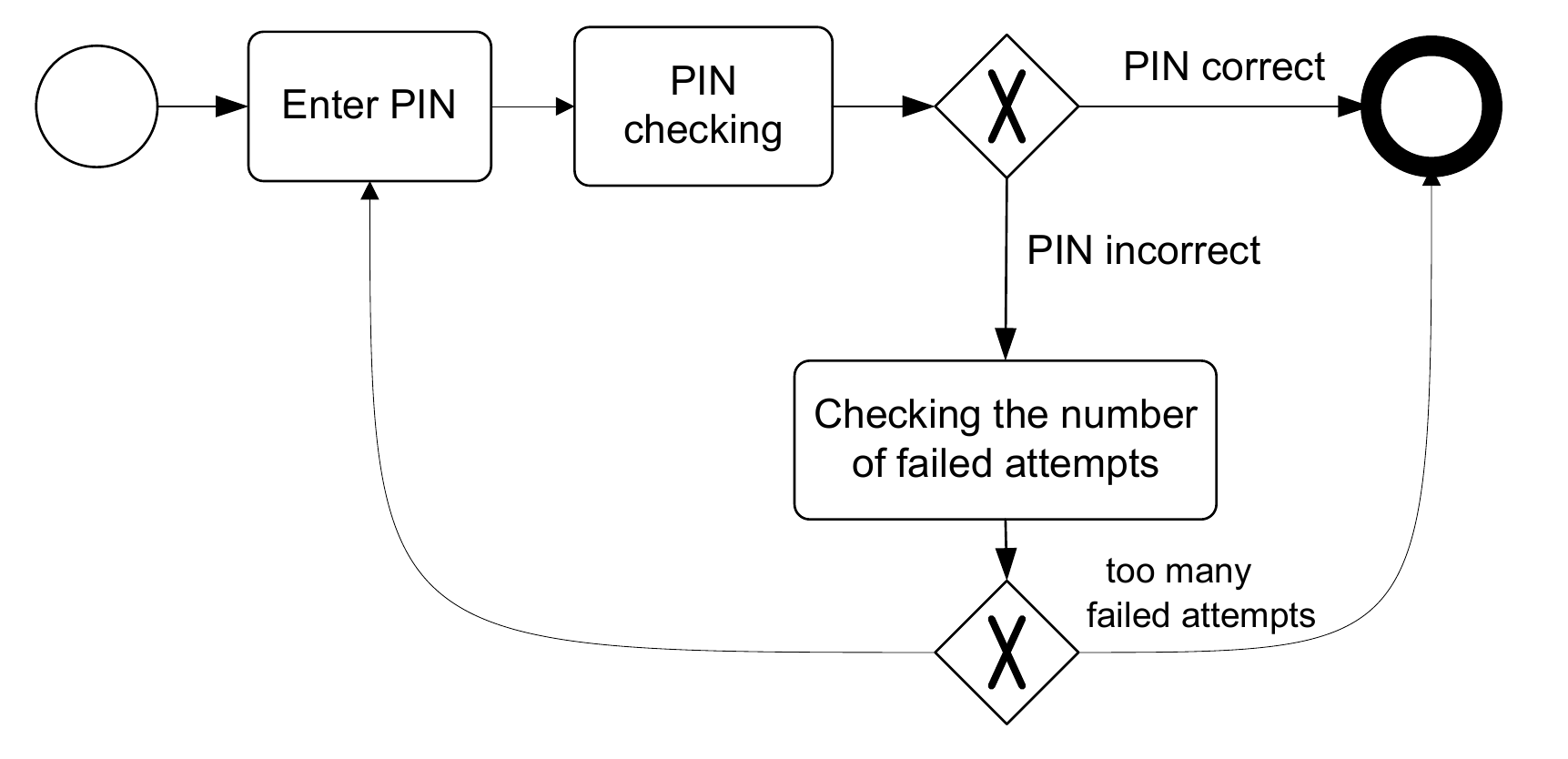}
\caption{Example of BPMN representation of cashpoint \textit{Authorization} subactivity}
\label{fig:cashpoint_table}
\vspace{-3mm}
\end{figure}

Several \textbf{integration scenarios} for XTT2 and BPMN are considered:
\begin{itemize}
\item \textbf{BPMN integration with XTT2}\\ This scenario assumes that BPMN and XTT2 have some intersecting parts, in which the integration of the two solutions can be performed. The general idea is as follows: BPMN is responsible for inference specification and hierarchization of the rulebase, and rule tables for some part of the system are designed in XTT2. 
Another example is a BPMN model of a cashpoint, shown in Fig.~\ref{fig:cashpoint_diagram} and~\ref{fig:cashpoint_table}.
\item \textbf{BPMN as a replacement of ARD+}\\ Because the abstraction level of ARD+ and BPMN seems to be similar, in this scenario BPMN is proposed to be used instead of present solution~-- ARD+. This assumes that mapping between BPMN tasks and XTT2 tables is one-to-one. A~prototype example of this approach is shown in Fig.~\ref{fig:visio_therm}. 
\item \textbf{BPMN representation of XTT2 table}\\ This is not a primary goal of integration. However, this could enable BPMN design of the whole XTT2 methodology, including single tables and rules. An example of this approach can be seen in Fig.~\ref{fig:tab_ms_bpmn}.
\end{itemize}

Because the assumed mapping in the first scenario may be not one-to-one, this scenario is highly complex. It requires well-prepared analysis and specification of both solutions as well as a detailed specification of the integration proposal. However, this is the best scenario for real-world cases. In the second one, in turn, the mapping is very simple, because each task is mapped to exactly one table. However, this solution does not provide the table schema, as it was in the case of ARD+. The third scenario is a rather academic one, because tables are already an efficient method of presenting rules, and their visual representation in another form may not be so useful~\cite{kluza2009csltr}.


\section{Conclusions and Future Work}
\label{sec:future}

The general problem considered in this paper is the RBS design and modularization.
The paper considers possible solutions to modularize rule bases with Drools, BPMN and XTT2.
However, these solutions have some limitations. 
The paper constitutes an overview and a proposal for integration of the three presented methodologies, 
which can be useful in solving the identified problems.

The work described here is partially in progress.
The rule translation from XTT2 to Drools is being developed and implemented.
The design is the result of the comparison of both semanticts (XTT2 and Drools) while the translation is achieved by the module to HQEd, writen in C++.
Drools 5 has some differences from its predecessor. The most important thing is that Drools Flow focuses more on a process management, rather than on the rule hierarchisation. In the previous version the main part was the block which refers to the rules in the knowledge base. In the new version there are much more blocks which provide strict integration with Java programming langauge.

Moreover, several issues concerning BPMN as an end-user notation are considered.
Future work will be focused on integration of the three described solutions.
The plan involves analysis of the BPMN notation for the purpose of Rule-Based Systems, 
which can be useful for implementation and application of the integrated methodology.
In a more distant future, the plan involves running selected BPMN models in the rule engine,
and comparison of the analysis of BPMN models via rule engine to executable BPEL4WS.


\bibliography{hekatebib/hekate-project,hekatebib/hekate-main,hekatebib/hekate-gjn,hekatebib/hekate-verification}

\end{document}